\documentclass{article}
\usepackage[final]{neurips_2023}
\usepackage[utf8]{inputenc} 
\usepackage[T1]{fontenc}    
\usepackage{hyperref}       
\usepackage{url}            
\usepackage{booktabs}       
\usepackage{amsfonts}       
\usepackage{nicefrac}       
\usepackage{microtype}      
\usepackage{xcolor}         
\usepackage{multirow}
\usepackage{graphicx}
\usepackage{amsmath}
\setcitestyle{square,numbers}

\title{Label Augmentation Method for Medical Landmark Detection in Hip Radiograph Images}
\author{%
  Yehyun Suh$^{1,2}$ \quad Peter Chan$^{3,4}$ \quad J. Ryan Martin$^{4}$ \quad Daniel Moyer$^{*1,2}$\\
  $^1$Vanderbilt Univserity $^2$Vanderbilt Institute for Surgery and Engineering \\
  $^3$University of Texas Southwestern Medical Center $^4$Vanderbilt University Medical Center \\
  \texttt{\{yehyun.suh, daniel.moyer\}@vanderbilt.edu} \\
}

\begin{document}
\maketitle
\begin{abstract}
    This work reports the empirical performance of an automated medical landmark detection method for predict clinical markers in hip radiograph images. Notably, the detection method was trained using a label-only augmentation scheme; our results indicate that this form of augmentation  outperforms traditional data augmentation and produces highly sample efficient estimators. We train a generic U-Net-based architecture under a curriculum consisting of two phases: initially relaxing the landmarking task by enlarging the label points to regions, then gradually eroding these label regions back to the base task. We measure the benefits of this approach on six datasets of radiographs with gold-standard expert annotations. \href{https://github.com/vine-lab-vu/Label-Augmentation}{Link to code.}
\end{abstract}

\section{Introduction}
    Total Hip Arthroplasty (THA), also known as Total Hip Replacement, is a standard procedure to address hip pain by removing and replacing the damaged joint with artificial components \cite{https://doi.org/10.1111/anae.15498}. Pre-surgical, intra-surgical, and post-surgical calculation of pelvic tilt, implant cup tilt, and evaluation of THA relies on medical landmarks on X-ray and Fluoroscope images of the patient's pelvis. In clinical practice, orthopedic clinicians manually select markers to make these assessments. Measurements of implant alignment during recovery and later regular use are vital as decision-making factors for potential future correction procedures, such as in the event of adverse implant positioning/alignment. Therefore, automating the process of medical landmark detection for orthopedic clinicians, especially in hip radiographs, is a target for computer vision in the medical field.
    
    In this work, we extend a label augmentation method, initially constructed for Total Knee Arthroplasty (TKA) \cite{suh2023dilationerosion}, to the THA case. We benchmark on both a knee dataset as well as five hip datasets with different imaging conditions. We show that our proposed method produces error distances on the order of $\sim$1-4 pixels, greatly outperforming baseline methods. Moreover, we show that traditional augmentation is actively harmful to this particular task due to the imaging protocol.
    
    
\section{Method}
    Conventional augmentation on the training images, such as rotating, flipping, resizing without padding, or color jittering distorts the patterns or creates invalid medical images, making the label of the data to be no longer preserved post-transformation. This is because standardized positioning of radiographs is critical for maintaining consistency, reproducibility, and accuracy. Therefore, instead of augmenting the training images, we implemented a Label Augmentation method that enhances the performance of the U-Net \cite{DBLP:journals/corr/RonnebergerFB15} using the same basic architecture and gradient-based learning. To be specific, the images' labels are first iterated a predetermined number of times to dilate them. It is allowed for these dilated labels to overlap. The dilated labels are used to train the prediction network (a U-Net), and after the training stages are completed, labels gradually erode over time. 
    
    As training goes on, we increase or decrease the size of each label, which causes label imbalance. Re-weighting the loss function, which biases predictions away from degenerate solutions, is a frequent static imbalance solution. We create a dynamic re-weighting method for dynamic imbalances, where $w$ is the label weight that would have been applied if we had not utilized the label augmentation and $\tilde{w}$ is the re-weighted $w$:
    \begin{align}
        \tilde{w} = w \times \frac{\text{input image size} - (\text{number of dilated pixels} + \text{number of label pixels}) }{(\text{number of dilated pixels} + \text{number of label pixels})}
    \end{align}
    By re-weighting labels by $\tilde{w}$, we keep the relative loss value of each label consistent across the learning curriculum. 

\section{Experiments \& Results}
    \begin{figure}
        \centering
        \includegraphics[width=0.28\textwidth]{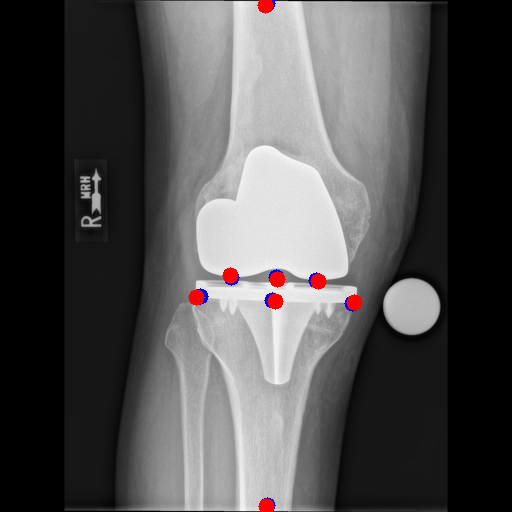}
        \includegraphics[width=0.28\textwidth]{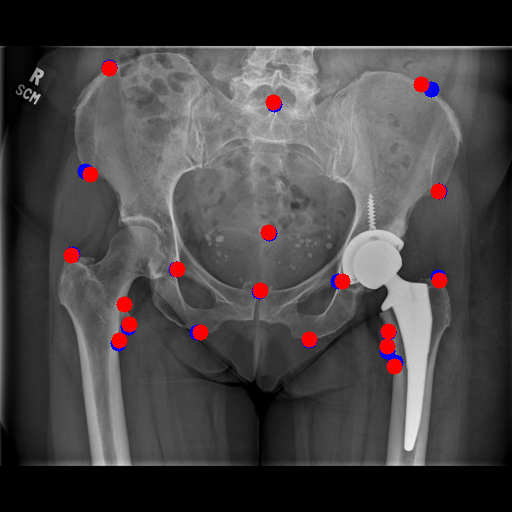}
        \includegraphics[width=0.28\textwidth]{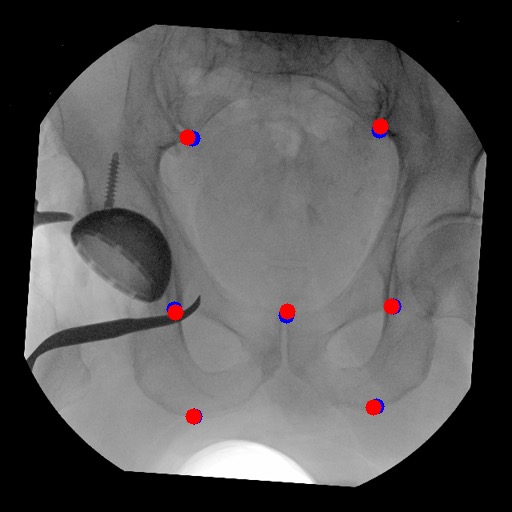} \\
        \includegraphics[width=0.28\textwidth]{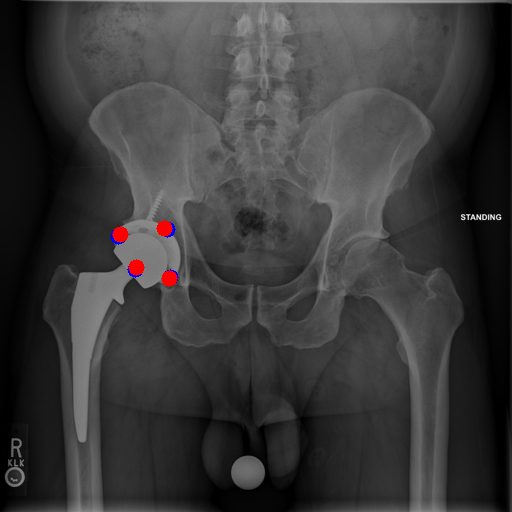}
        \includegraphics[width=0.28\textwidth]{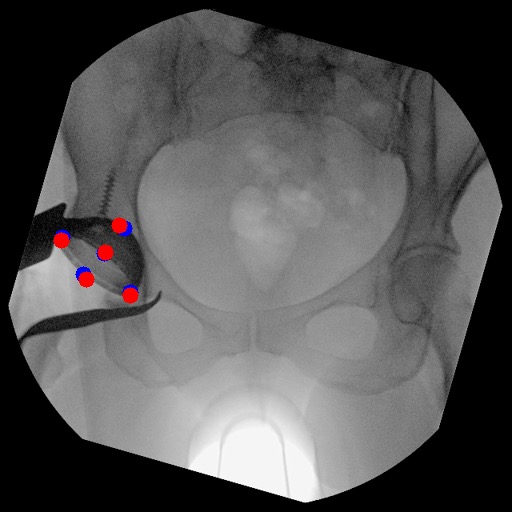}
        \includegraphics[width=0.28\textwidth]{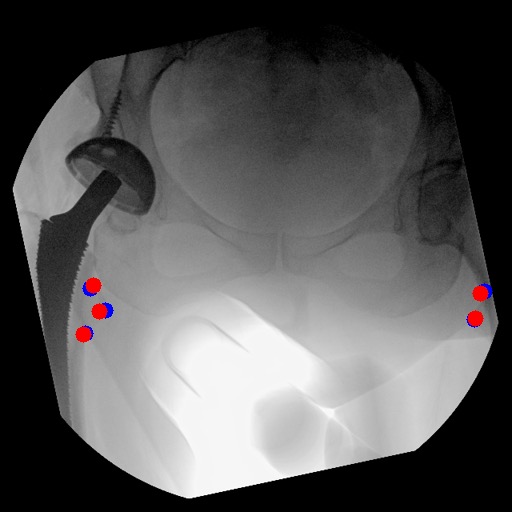}

        \caption{From left to right and top to bottom, each image represents an X-ray of the knee, an X-ray of the pelvis, a Fluoroscope of the pelvis, an X-ray of the pelvis with hip implant, a Fluorocope of the pelvis with hip implant, and a Fluoroscope of the pelvis with trochanter with predicted outputs (red) and ground truth labels (blue).}
        \label{fig:predict-label}
    \end{figure}

    \begin{table}
    \centering
    \caption{Radio. stands for radiographs, Fluoro indicates fluoroscopes, Aug is the training dataset augmentation, and SDR represents the Success Detection Rate.}
    \label{internal}
        \begin{tabular}{|c|cccccc|}
        \hline
        \multirow{2}{*}{Dataset} & \multicolumn{6}{c|}{Validation Set (SDR(\%))} \\ \cline{2-7} 
         & \multicolumn{1}{c|}{Experiment} & \multicolumn{1}{c|}{Mean RMSE} & \multicolumn{1}{c|}{< 2} & \multicolumn{1}{c|}{< 2.5} & \multicolumn{1}{c|}{< 3} & < 4 \\ \hline
        \multirow{4}{*}{Knee Radio.} & \multicolumn{1}{c|}{Baseline} & \multicolumn{1}{c|}{124.82} & \multicolumn{1}{c|}{0.29} & \multicolumn{1}{c|}{0.29} & \multicolumn{1}{c|}{0.58} & 0.58 \\ \cline{2-7} 
         & \multicolumn{1}{c|}{Baseline + Aug} & \multicolumn{1}{c|}{137.08} & \multicolumn{1}{c|}{0} & \multicolumn{1}{c|}{0} & \multicolumn{1}{c|}{0} & 0 \\ \cline{2-7} 
         & \multicolumn{1}{c|}{Proposed + Aug} & \multicolumn{1}{c|}{5.26} & \multicolumn{1}{c|}{39.53} & \multicolumn{1}{c|}{54.36} & \multicolumn{1}{c|}{63.95} & 68.60 \\ \cline{2-7} 
         & \multicolumn{1}{c|}{\textbf{Proposed}} & \multicolumn{1}{c|}{\textbf{1.79}} & \multicolumn{1}{c|}{\textbf{63.08}} & \multicolumn{1}{c|}{\textbf{78.49}} & \multicolumn{1}{c|}{\textbf{88.37}} & \textbf{95.64} \\ \hline
        \multirow{4}{*}{Pelvis Radio.} & \multicolumn{1}{c|}{Baseline} & \multicolumn{1}{c|}{175.03} & \multicolumn{1}{c|}{0} & \multicolumn{1}{c|}{0} & \multicolumn{1}{c|}{0} & 0 \\ \cline{2-7} 
         & \multicolumn{1}{c|}{Baseline + Aug} & \multicolumn{1}{c|}{180.77} & \multicolumn{1}{c|}{0} & \multicolumn{1}{c|}{0} & \multicolumn{1}{c|}{0.08} & 0.15 \\ \cline{2-7} 
         & \multicolumn{1}{c|}{Proposed + Aug} & \multicolumn{1}{c|}{8.75} & \multicolumn{1}{c|}{17.54} & \multicolumn{1}{c|}{26.53} & \multicolumn{1}{c|}{33.33} & 41.87 \\ \cline{2-7} 
         & \multicolumn{1}{c|}{\textbf{Proposed}} & \multicolumn{1}{c|}{\textbf{3.98}} & \multicolumn{1}{c|}{\textbf{38.15}} & \multicolumn{1}{c|}{\textbf{52.08}} & \multicolumn{1}{c|}{\textbf{63.38}} & \textbf{74.15} \\ \hline
        \multirow{4}{*}{Pelvis Fluoro.} & \multicolumn{1}{c|}{Baseline} & \multicolumn{1}{c|}{129.70} & \multicolumn{1}{c|}{0} & \multicolumn{1}{c|}{0} & \multicolumn{1}{c|}{0} & 0 \\ \cline{2-7} 
         & \multicolumn{1}{c|}{Baseline + Aug} & \multicolumn{1}{c|}{135.50} & \multicolumn{1}{c|}{0} & \multicolumn{1}{c|}{0} & \multicolumn{1}{c|}{0} & 0 \\ \cline{2-7} 
         & \multicolumn{1}{c|}{Proposed + Aug} & \multicolumn{1}{c|}{8.74} & \multicolumn{1}{c|}{15.27} & \multicolumn{1}{c|}{22.18} & \multicolumn{1}{c|}{27.29} & 36.45 \\ \cline{2-7} 
         & \multicolumn{1}{c|}{\textbf{Proposed}} & \multicolumn{1}{c|}{\textbf{3.09}} & \multicolumn{1}{c|}{\textbf{41.67}} & \multicolumn{1}{c|}{\textbf{55.92}} & \multicolumn{1}{c|}{\textbf{65.13}} & \textbf{74.78} \\ \hline
        \multirow{4}{*}{Pelvis Radio. w Implant} & \multicolumn{1}{c|}{Baseline} & \multicolumn{1}{c|}{81.08} & \multicolumn{1}{c|}{0} & \multicolumn{1}{c|}{0} & \multicolumn{1}{c|}{0} & 0 \\ \cline{2-7} 
         & \multicolumn{1}{c|}{Baseline + Aug} & \multicolumn{1}{c|}{151.79} & \multicolumn{1}{c|}{0} & \multicolumn{1}{c|}{0} & \multicolumn{1}{c|}{0} & 0 \\ \cline{2-7} 
         & \multicolumn{1}{c|}{Proposed + Aug} & \multicolumn{1}{c|}{5.88} & \multicolumn{1}{c|}{24.04} & \multicolumn{1}{c|}{42.26} & \multicolumn{1}{c|}{54.76} & 66.07 \\ \cline{2-7} 
         & \multicolumn{1}{c|}{\textbf{Proposed}} & \multicolumn{1}{c|}{\textbf{1.82}} & \multicolumn{1}{c|}{\textbf{64.35}} & \multicolumn{1}{c|}{\textbf{80.10}} & \multicolumn{1}{c|}{\textbf{86.65}} & \textbf{93.52} \\ \hline
        \multirow{4}{*}{Pelvis Fluoro. w Implant} & \multicolumn{1}{c|}{Baseline} & \multicolumn{1}{c|}{154.86} & \multicolumn{1}{c|}{0} & \multicolumn{1}{c|}{0} & \multicolumn{1}{c|}{0} & 0 \\ \cline{2-7} 
         & \multicolumn{1}{c|}{Baseline + Aug} & \multicolumn{1}{c|}{150.35} & \multicolumn{1}{c|}{0} & \multicolumn{1}{c|}{0} & \multicolumn{1}{c|}{0} & 0 \\ \cline{2-7} 
         & \multicolumn{1}{c|}{Proposed + Aug} & \multicolumn{1}{c|}{5.46} & \multicolumn{1}{c|}{24.14} & \multicolumn{1}{c|}{45.98} & \multicolumn{1}{c|}{56.32} & 64.37 \\ \cline{2-7} 
         & \multicolumn{1}{c|}{\textbf{Proposed}} & \multicolumn{1}{c|}{\textbf{2.08}} & \multicolumn{1}{c|}{\textbf{47.06}} & \multicolumn{1}{c|}{\textbf{70.59}} & \multicolumn{1}{c|}{\textbf{81.18}} & \textbf{95.29} \\ \hline
        \multirow{4}{*}{Pelvis Fluoro. w Trochanter} & \multicolumn{1}{c|}{Baseline} & \multicolumn{1}{c|}{151.89} & \multicolumn{1}{c|}{0} & \multicolumn{1}{c|}{0} & \multicolumn{1}{c|}{0.47} & 1.42 \\ \cline{2-7} 
         & \multicolumn{1}{c|}{Baseline + Aug} & \multicolumn{1}{c|}{166.79} & \multicolumn{1}{c|}{0} & \multicolumn{1}{c|}{0} & \multicolumn{1}{c|}{0} & 0 \\ \cline{2-7} 
         & \multicolumn{1}{c|}{Proposed + Aug} & \multicolumn{1}{c|}{6.46} & \multicolumn{1}{c|}{17.45} & \multicolumn{1}{c|}{25.0} & \multicolumn{1}{c|}{29.72} & 36.79 \\ \cline{2-7} 
         & \multicolumn{1}{c|}{\textbf{Proposed}} & \multicolumn{1}{c|}{\textbf{3.55}} & \multicolumn{1}{c|}{\textbf{34.90}} & \multicolumn{1}{c|}{\textbf{43.75}} & \multicolumn{1}{c|}{\textbf{54.17}} & \textbf{66.67} \\ \hline
        \end{tabular}
    \end{table}

    \begin{table}
    \centering
    \caption{Label 1 to 5 stands for Teardrops, most inferior aspect of Ischium, medial most point on Lesser Trochanter, superior most point on Greater Trochanter, and center of Sacrococcygeal Juntion.}
    \label{mayo}
        \begin{tabular}{|c|cccccc|}
        \hline
        \multirow{2}{*}{Model} & \multicolumn{6}{c|}{RMSE(mm)} \\ \cline{2-7} 
         & \multicolumn{1}{c|}{Mean} & \multicolumn{1}{c|}{Label 1} & \multicolumn{1}{c|}{Label 2} & \multicolumn{1}{c|}{Label 3} & \multicolumn{1}{c|}{Label 4} & \multicolumn{1}{l|}{Label 5} \\ \hline
        Mulford et al. \cite{MULFORD2023} & \multicolumn{1}{c|}{\textbf{3.3}} & \multicolumn{1}{c|}{2.7} & \multicolumn{1}{c|}{\textbf{3.1}} & \multicolumn{1}{c|}{\textbf{2.1}} & \multicolumn{1}{c|}{3.0} & \textbf{5.6} \\ \hline
        Ours & \multicolumn{1}{c|}{3.60} & \multicolumn{1}{c|}{\textbf{2.19}} & \multicolumn{1}{c|}{3.43} & \multicolumn{1}{c|}{2.13} & \multicolumn{1}{c|}{\textbf{2.06}} & 8.17 \\ \hline
        \end{tabular}
    \end{table}

    To evaluate our model, we have used six datasets: an X-ray of the knee (216 images, 8 lables), an X-ray/Fluoroscope of the pelvis (329/286 images, 20/8 labels), an X-ray/Fluoroscope of the pelvis that consists implant (210/84 images, 4/5 labels), Fluoroscope of the pelvis that consists trochanter (159 images/ 6 lables). Internal datasets were annotated by one medical student and were divided 4:1 for the train:validation. Before resizing, all the images and the corresponding label masks were padded to have the same height and width. They were resized to a standard size (512$\times$512) and normalized to [0,1]. 
    
    Our model's encoder used pretrained weight from ImageNet \cite{5206848} and used the Label Augmentation method and re-weighting scheme with pixel-wise cross-entropy loss to train the model. For the label augmentation, we have dilated the pixel for 65 iterations and eroded by 10 iterations every 50 epochs. As a validation metric, we utilize the Root Mean Squared Error (RMSE) of the pixel-wise distance between the most probable pixel (the maximal output logit value) and the groundtruth label position. 

    We compared the performance of our method to three different methods: a baseline U-Net, U-net with training set augmentation, and U-net with both training set augmentation and label augmentation. As shown in Table~\ref{internal}, in comparison to all baseline methods, our training method performs better: the mean RMSE across datasets has decreased from over 100 to less than 4. Also, we have done random rotation of a maximum of 20 degrees for the training set augmentation, and it did not work on both baseline U-net and U-net with label augmentation. We plot the results in Figure~\ref{fig:predict-label}. 
    
    We also compared our results with Mulford et al. \cite{MULFORD2023} on pelvis X-ray as shown in Table~\ref{mayo}. Despite our model having one-third of the dataset size, we have outperformed Mulford et al. \cite{MULFORD2023} in some labels where they had 1000 images for training and validation. Also, considering that our dataset had 55 images that did not have the center of the Sacrococcygeal Junction (Label 5) and only one annotator, our model has shown a decent performance.

\section{Conclusion \& Discussion}
    Our study presents a promising approach to automated medical landmark detection in hip radiographs using the Label Augmentation method combined with dynamic re-weighting. We have shown that the conventional augmentation on medical dataset decreases the performance of the model and have shown the potential to be used in general landmark detection in medical imaging. However, our results were based on the validation set not test set and our model has yet to show outstanding performance on some of the tasks in datasets such as on X-ray of the pelvis. We hypothesize that this may be due to the inconsistency in data labeling since we had only one annotator compared to  Mulford et al. \cite{MULFORD2023}, where they had two annotators and selected the medial points from each annotator. In our future works, we plan to collect prospective test dataset and experiment inter-rater reliability to test the performance of our model to determine the noise ceiling of prediction accuracy and implement this work in diverse fields of medicine.
    
\section*{Potential Negative Societal Impact}
    Since the model is trained on data collected from actual patients, if this data is not representative of the entire population, the automated systems may be biased against certain groups of people, leading to disparities in healthcare access and outcomes.
\section*{Acknowledgement}

This work was funded in part by NSF 2321684 and the Wellcome LEAP Multi-Channel Psych program, as well as a seed grant from the Vanderbilt Institute for Surgery and Engineering (VISE).
    
\medskip
\bibliography{reference}

\end{document}